\documentclass[journal,10pt]{IEEEtran}
\usepackage{graphicx} 
\usepackage{algorithm}
\usepackage{algorithm}
\usepackage{algorithmic}
\usepackage{multirow}
\usepackage{amsmath}
\usepackage{amsfonts}
\usepackage{pdflscape}
\usepackage{subfiles} 
\usepackage{booktabs}
\usepackage{url}
\graphicspath{{Images/}}

\DeclareMathOperator*{\argmax}{arg\,max}

\title{HD-CB: The First Exploration of Hyperdimensional Computing for Contextual Bandits Problems}
    \author{
        \IEEEauthorblockN{
            Marco Angioli,
            Antonello Rosato,
            Marcello Barbirotta,
            Rocco Martino,
            Francesco Menichelli, 
            Mauro Olivieri\\
        }
        \IEEEauthorblockA{Dept. of Information Engineering, Electronics and Telecommunications,\\
        Sapienza University of Rome, Italy, \\
        Email: \{name.surname\}@uniroma1.it\\
        }
    }

\begin{document}

\maketitle
\begin{abstract}
Hyperdimensional Computing (HDC), also known as Vector Symbolic Architectures, is a computing paradigm that combines the strengths of symbolic reasoning with the efficiency and scalability of distributed connectionist models in artificial intelligence.
HDC has recently emerged as a promising alternative for performing learning tasks in resource-constrained environments thanks to its energy and computational efficiency, inherent parallelism, and resilience to noise and hardware faults.

This work introduces the Hyperdimensional Contextual Bandits (HD-CB): the first exploration of HDC to model and automate sequential decision-making Contextual Bandits (CB) problems. The proposed approach maps environmental states in a high-dimensional space and represents each action with dedicated hypervectors (HVs). At each iteration, these HVs are used to select the optimal action for the given context and are updated based on the received reward, replacing computationally expensive ridge regression procedures required by traditional linear CB algorithms with simple, highly parallel vector operations.
We propose four HD-CB variants, demonstrating their flexibility in implementing different exploration strategies, as well as techniques to reduce memory overhead and the number of hyperparameters. 
Extensive simulations on synthetic datasets and a real-world benchmark reveal that HD-CB consistently achieves competitive or superior performance compared to traditional linear CB algorithms, while offering faster convergence time, lower computational complexity, improved scalability, and high parallelism.
\end{abstract}

    
        
\section{Introduction}

\IEEEPARstart{M}{}any real-world applications require sequential decision-making, where the goal is to select the optimal action from a finite set of alternatives based on the observed context or state of the environment \cite{CB_review}. 
These tasks can be formally modelled as Contextual Bandits (CB) problems, where an agent iteratively interacts with the environment, observes the context, selects an action, and receives a reward related to the chosen option. This reward updates the model parameters and guides the agent in refining its decision-making process, enabling it to adapt and improve through continuous interaction with the environment. The goal of the agent is to maximize cumulative reward over time by balancing exploring less-known actions with exploiting previously gathered knowledge \cite{li2010contextual}. 
CB problems are prevalent across various domains, including online advertising \cite{li2010contextual}, personalized recommendations \cite{Netflix, MICROSOFT, NYT, GAME}, hardware reconfiguration \cite{angioli_automatic}, edge computing \cite{EDGE1, EDGE2}, and medical decision-making \cite{MD1, MD2}.

In this work, we propose a novel, lightweight, and highly parallel approach for modelling and automating CB problems using Hyperdimensional Computing (HDC).
HDC is a computing paradigm that mimics the cognitive processes of the human brain by mapping concepts to high-dimensional distributed representations called hypervectors (HVs), which are combined and compared using simple vector arithmetic. 
In recent years, HDC has found applications across various fields, demonstrating concrete advantages in computational and energy efficiency, scalability, convergence time, resilience, and parallelism \cite{Kenny, RW_review_1, RW_review_2, RW_review_3}.

The approach presented in this paper, referred to as Hyperdimensional Contextual Bandits (HD-CB), builds on the mathematical foundation of HDC to generate representative vectors in the high-dimensional space that compactly encapsulate information about the expected quality of each action in a given context.
During the online decision-making process, the current state is mapped into the high-dimensional space and compared with the representative HV of each action to determine the one with the highest expected reward.
After the chosen action is executed and a reward is received, the model updates the corresponding action HV, moving it closer to or further away from the encoded context to better reflect the expected quality of the action.

The main contributions of the proposed study are the following:
\begin{itemize}
\item The work represents the first exploration of HDC to model and automate CB problems.
\item It details the proposed approach in four variants tailored to different exploration strategies and resource constraints.
\item It analyzes the performance of HD-CB in terms of average reward, execution time and convergence speed while varying the problem parameters and the HV size.
\item It compares all of the proposed schemes with traditional CB algorithms from the literature using synthetic datasets and real benchmarks.
\end{itemize}


        
\section{Background and Related Works}\label{Background_and_Relatedworks}

\subsection{Contextual Bandits}
\begin{figure}[!h]
        \centering
        \includegraphics[width=0.7\columnwidth]{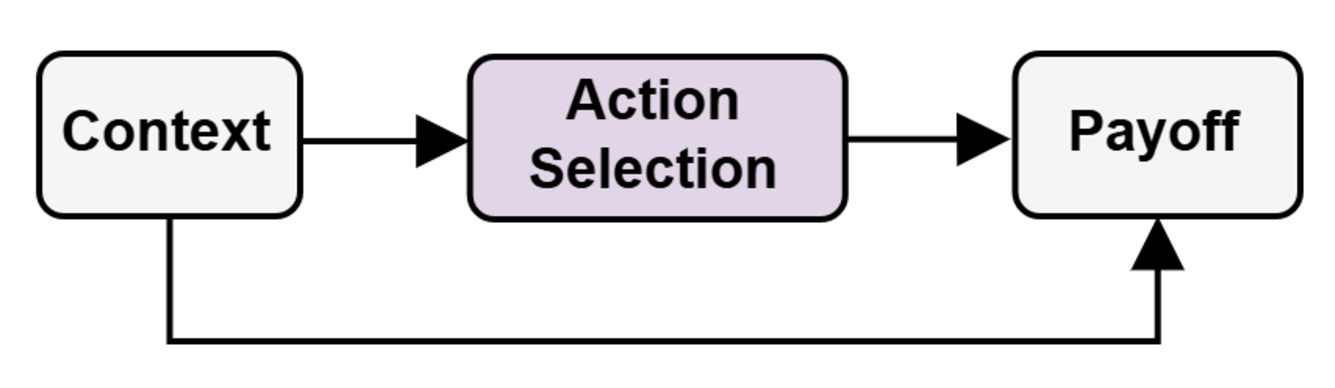}
        \caption{Schematic representation of contextual bandits problems. At each iteration, the algorithms observe a context, pick an action and receive a reward related to the chosen option.}
        \label{fig:contextual_bandits_scheme}
    \end{figure}

    Contextual Multi-Armed Bandits, or simply Contextual Bandits, are a class of sequential decision-making problems where an agent interacts with the environment to learn and adapt in real-time through a reward-driven process \cite{CB_review}.
    Given a finite set of \( N \) possible actions (or ``arms``), at each time step \( t \), the agent interacts with the environment observing a context vector \( x_{t,a} \in \mathbb{R}^d \) for each action $a$, selects an action \( a_t \), and receives a reward (or payoff) \( r_{t,a_t} \), as depicted in Figure \ref{fig:contextual_bandits_scheme}. This reward is then used to update the model parameters, enabling it to adapt and gather information to predict the relationship between the observed context and the reward associated with each action \cite{angioli_automatic}.

     A widely studied branch of algorithms for modelling CB problems, which will be considered as a reference for comparison in this work, assumes a linear relationship between the reward and the context \cite{li2010contextual, li2011unbiased}.
     These algorithms analytically model each action $a$ through a $d \times d$ matrix $\mathbf{A}_a$ and a $d \times 1$ vector $\mathbf{b}_a$. At each time step $t$, the algorithm employs the ridge regression described in (\ref{eq:eg_disj}) to estimate the expected reward ($p_{t, a}$) for each action $a$ as a function of the current context.
    
    \begin{equation}\label{eq:eg_disj}
            p_{t, a} = (\mathbf{A}_{a}^{-1} \mathbf{b}_{a})^{\top} \mathbf{x}_{t, a}
    \end{equation}

    After computing the expected reward, the algorithm employs exploration strategies to select an action, balancing exploring less-known options with exploiting previously gathered information, ensuring the model can adapt and improve over time while avoiding premature convergence to suboptimal solutions.
    Two widely used strategies in CB algorithms are the epsilon-greedy and the Upper Confidence Bound (UCB), which, when integrated into linear algorithms, are referred to as LinearEPS and LinearUCB, respectively \cite{CB_review, li2010contextual, zr_obp}. In the epsilon-greedy technique, at each time step, with a probability of $1 - \epsilon$, the model selects the action with the highest estimated reward, while in the remaining cases, it explores by selecting a random action. 
    This strategy ensures continuous exploration over time but does not take into account the confidence of the model, potentially resulting in inefficient choices.
    
    The UCB exploration strategy \cite{li2010contextual}, on the other hand, incorporates a second factor into the reward estimation as shown in (\ref{eq:ucb_disj}).
        \begin{equation}\label{eq:ucb_disj}
        p_{t, a} =  (\mathbf{A}_{a}^{-1} \mathbf{b}_{a})^{\top} \mathbf{x}_{t, a} + \alpha \sqrt{\mathbf{x}_{t, a}^{\top} \mathbf{A}_{a}^{-1} \mathbf{x}_{t, a}}
    \end{equation}

    This term ( $ \sqrt{\mathbf{x}_{t, a}^{\top} \mathbf{A}_{a}^{-1} \mathbf{x}_{t, a}}$), modulated by the hyperparameter $\alpha \in [0,1]$, represents the confidence interval for the predicted payoff and thus accounts for the uncertainty of the model in the reward estimation, dynamically balancing exploration and exploitation effectively.
    
    Regardless of the exploration strategy employed, once an action is selected, the environment yields a reward ($r_{t,a}$) to the model, which is used to update the matrices associated with the chosen action, as shown in (\ref{eq:update_disj}).
    
    \begin{equation}\label{eq:update_disj}
         \begin{array}{l}
            \mathbf{A}_{a_{t}} = \mathbf{A}_{a_{t}}+\mathbf{x}_{t, a_{t}} \mathbf{x}_{t, a_{t}}^{\top}
            \\
            \mathbf{b}_{a_{t}} = \mathbf{b}_{a_{t}}+r_{t} \mathbf{x}_{t, a_{t}}
        \end{array}
    \end{equation}
 
    \subsection{Hyperdimensional Computing}
        HDC is a computing paradigm inspired by the intricate workings of the human brain. In HDC, information is encoded using very long vectors, typically in the order of thousands of dimensions, akin to kernel expansion methods in machine learning. These HVs distribute information across their elements, encoding data through the combination of active components, mimicking the distributed neural activity patterns observed in the brain \cite{RW_review_3}. 

        At the core of HDC is the principle that fundamental concepts, symbols, or elements of a problem can be mapped to random HVs that result nearly orthogonal—i.e., linearly independent—thanks to the mathematical properties of high-dimensional spaces. HDC models manipulate and combine these HVs using three primitive arithmetic operations—bundling, binding, and permutation—and a comparison one, the similarity. Together, these operations form the mathematical foundation of HDC and enable the composition of complex information without increasing dimensionality.
        The \emph{Bundling} ($\oplus$) combines two HVs into one that retains the key characteristics of both inputs, mimicking the concurrent activation of multiple neural patterns. Binding ($\otimes$) associates two different concepts, creating a new vector representing their combined relationship while remaining dissimilar to the originals; this operation is associative, commutative, self-invertible, and can distribute over bundling. 
        The \emph{Permutation} operation ($\rho$) reorders the elements of an HV, producing a dissimilar vector. This operation is invertible and can be distributed over both binding and bundling. 
        Finally, the \emph{similarity} measurement ($\delta$) assesses the angular distance between HVs in high-dimensional space. HVs representing related concepts will be closer in the space, with a smaller angular distance indicating higher similarity.
        The implementation of these operations varies depending on the characteristics of the hyperspace in which the HVs are defined \cite{schlegel2022comparison}. The elements of HVs, in fact, can adopt different \emph{data types}, such as binary, bipolar, ternary, quantized, integer, real, or complex \cite{RW_review_4}, to balance performance and computational complexity \cite{Kenny}. 
        
\begin{figure}[!t]
            \centering
            \includegraphics[width=0.6\columnwidth]{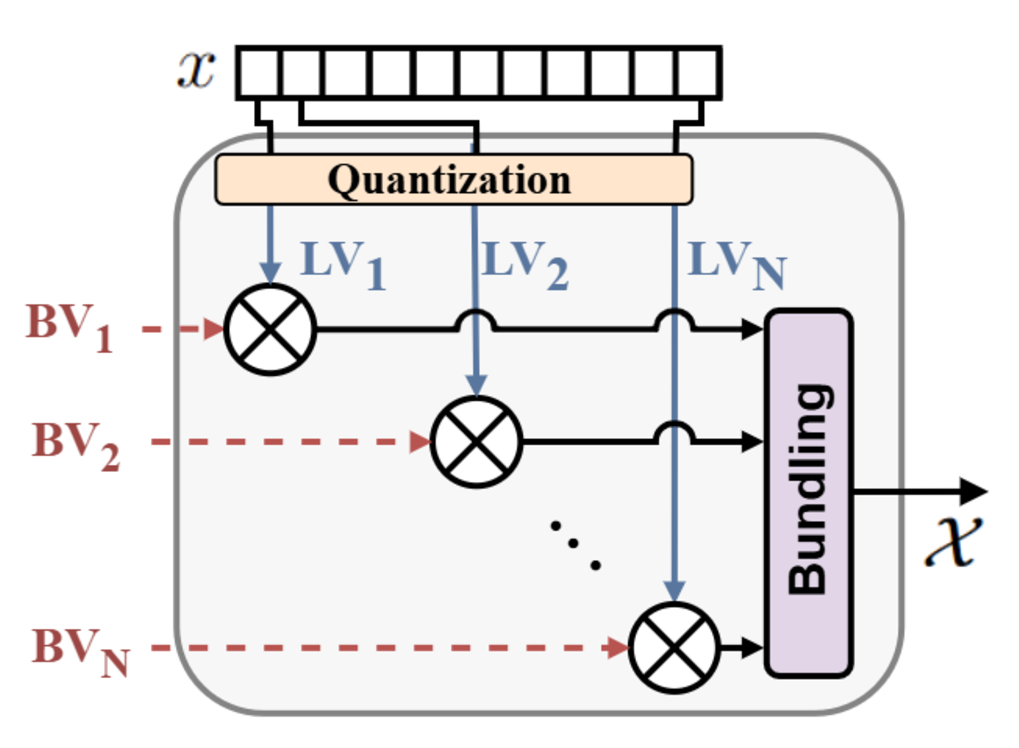}
            \caption{Schematic representation of the encoding unit. Each feature index of $x$ is represented by a dedicated random base-vector, while the corresponding values are discretized and mapped to level-vectors. The binding creates feature id-value pairs, while bundling produces the final encoded HV, $\mathcal{X}$ }
            \label{fig: encoding_unit}
        \end{figure}

        These arithmetic operations enable HDC to project diverse objects and their relationships into high-dimensional space, preserving semantic properties.
        This process, known as \emph{encoding}, is performed in this work using the \emph{record-based} encoding technique, depicted in Figure \ref{fig: encoding_unit}.
        In this technique, each feature-ID or index \(i\) of the input feature vector \(\mathbf{x} = [x_1, x_2, \dots, x_N]\) is mapped to a randomly generated base vector (\(BV_i\)).  This base vector is then bound to a level vector (\(LV_i\)), derived by discretizing the corresponding feature value and mapping it into the high-dimensional space, ensuring that similar scalar values in the original space are projected into similar HVs.
        This binding creates the features ID-value pairs, which are then aggregated through the bundling operation, resulting in the final encoded HV, \(\mathcal{X}\).

        \subsection{Related Works}

        While CB problems and HDC have been extensively studied independently, this work, to the best of our knowledge, marks the first attempt to leverage HDC for modeling and automating CB problems.
        
        Over the past decade, CB algorithms have been gaining in popularity due to their effectiveness and flexibility in solving sequential decision problems across various fields. For a comprehensive and up-to-date review of most of the applications in the field, the reader can refer to \cite{CB_review}.  In healthcare, these algorithms have been applied to improve treatment allocation in clinical trials \cite{MD1} and propose personalized dosing strategies \cite{MD2}. 
        In Microsoft \cite{MICROSOFT}, The New York Times \cite{NYT} and Yahoo \cite{li2010contextual} recommendation systems, CB algorithms have been widely utilized to enhance personalized content and news article delivery while Netflix \cite{Netflix} uses them to improve movie recommendations, demonstrating the ability to learn user preferences over time, enhancing the experience and engagement. 
        In dialogue systems, CB has been demonstrated to improve response selection in conversational agents, leading to more natural and effective user interactions \cite{liu2018customized}.
        Furthermore, recent advancements have shown the potential of CB algorithms in enhancing hardware efficiency by automatically reconfiguring the architecture at run-time according to the current workload \cite{angioli_automatic}.
        These application scenarios, spanning a broad and varied scope, underscore the importance of continued research into CB algorithms. The diverse challenges and environments in which CB algorithms are employed highlight the need to develop new techniques to enhance their accuracy, reduce execution and convergence times, lower computational demands, and minimize energy consumption. For these reasons, in this work, we explore HDC as a novel approach to address and automate these problems.

        HDC has gained significant attention for its ability to tackle reasoning problems across various domains, including robotics \cite{RW_robotic}, healthcare \cite{RW_Healthcare_1}, natural language processing \cite{RW_Textclassification_1}, speech recognition \cite{RW_Speech} and reservoir computing \cite{Rosato_RHDC}, for modelling and solving classification \cite{RW_classification_review}, clustering \cite{HDCluster} and regression \cite{RegHD} tasks. 
        In these applications, HDC has demonstrated performance comparable to state-of-the-art AI approaches while offering highly parallel, robust, fast, low-power, and lightweight solutions \cite{RW_review_1, RW_review_2, RW_review_3}, making it an ideal choice for cognitive tasks in resource-constrained environments \cite{RW_review_1, RW_review_4} and for hardware implementations \cite{RW_DNA, AeneasHDC_2024, HDCU}. 
        Recently, HDC has also started gaining attention in Reinforcement Learning (RL). Authors in \cite{ni2022qhd} introduce QHD, a novel algorithm that uses HDC to model Q-learning, achieving improved performance and up to a 34.6× speedup in real-time applications compared to Deep Q-Networks. Similarly, the work in \cite{ni2022hdpg} presents HDPG, a hyperdimensional policy-based reinforcement learning algorithm for continuous control tasks in robotics, offering superior performance and faster execution compared to Deep Neural Network-based RL methods.


    \section{Hyperdimensional Contextual Bandits (HD-CB)}\label{ProposedModels}
    This work introduces an innovative approach for modelling and automating online sequential decision-making problems using HDC. By leveraging the mathematical properties of high-dimensional spaces we develop a new family of models called Hyperdimensional Contextual Bandits (HD-CB) that replace
    computationally expensive ridge regression procedures in traditional CB algorithms with simple, highly parallel vector operations.
    This section represents the core of the paper and is organized into multiple subsections to highlight the development and significance of each variant in the proposed framework. Section \ref{HDCB_basics} establishes the foundation of HD-CB, detailing how actions and context are encoded into the high-dimensional space, how these representations are utilized to estimate the expected reward, and how the model is dynamically updated.
    Building on this foundation, Section \ref{HDCB_eps} introduces the first formulation of the HD-CB algorithm, which utilizes an $\epsilon$-greedy exploration strategy.
    Section \ref{HDCB_unc} extends the framework by proposing an uncertainty-driven variant that incorporates a confidence interval to refine the expected payoff estimation for each action.
    Sections \ref{HDCB_unc2} and \ref{HDCB_unc3} present alternative formulations of the uncertainty-driven algorithm aimed to simplify the model by eliminating one hyperparameter and reducing memory requirements, respectively, making the approach more efficient and adaptable to resource-constrained environments.

    
\subsection{Basics}\label{HDCB_basics}
\begin{figure}[!t]
        \centering
        \includegraphics[width=1\columnwidth]{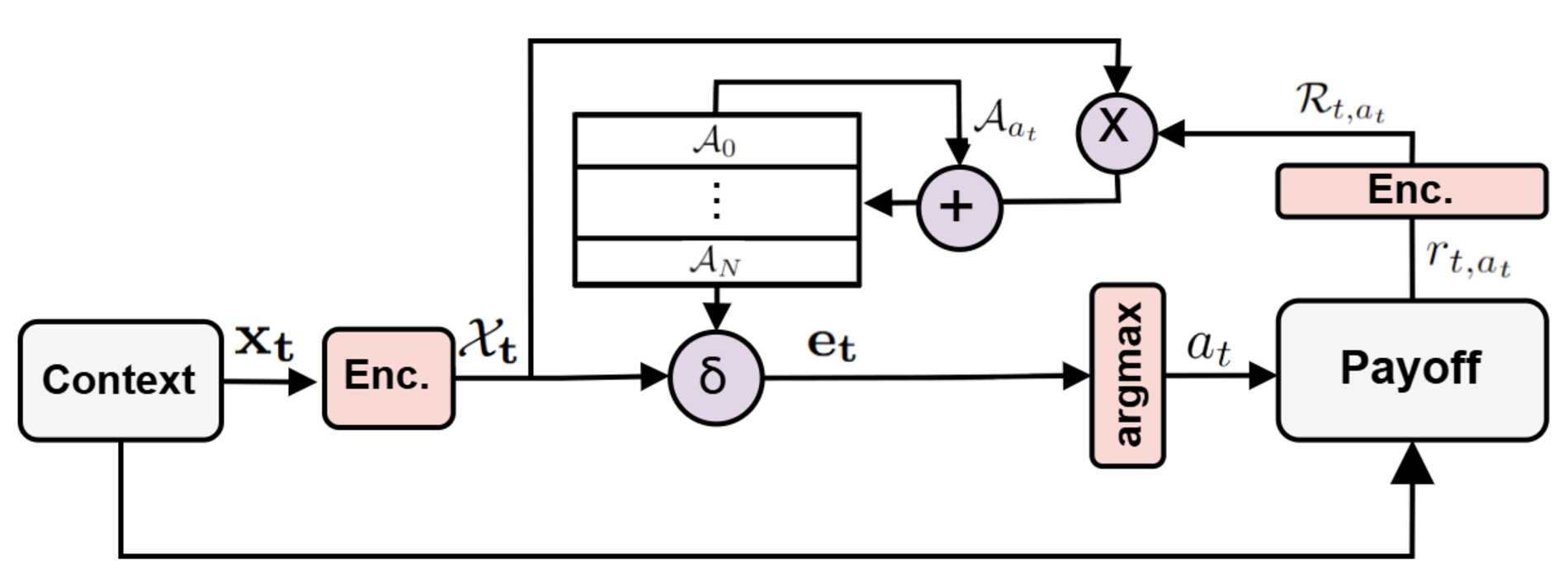}
        \caption{Schematic illustration of the fundamental working principle of HD-CB. The collection of encoded hypervectors $\mathcal{X}_{t,a}$ for all actions $a$ at time step $t$ is denoted as $\mathbf{\mathcal{X}_{t}}$, while the set of estimated payoffs $e_{t,a}$ for all actions is represented as $\mathbf{e_{t}}$.}
        \label{fig:HD-CB_basics}
    \end{figure}

    Figure \ref{fig:HD-CB_basics} provides an overview of the fundamental working principle of the proposed HD-CB algorithm. Each action $a$ of the possible $N$ is modeled by a dedicated representative vector $\mathcal{A}_a$ in the high-dimensional space, which is initialized to zero and iteratively updated to capture the expected quality of the action within the current context or state of the environment.

    At each iteration $t$, the model maps the observed context \(x_{t,a}\) into a high-dimensional HV \(\mathcal{X}_{t,a}\) using the \emph{record-based} encoding technique introduced in Section \ref{Background_and_Relatedworks}.

    The estimated payoff \(e_{t,a}\) of each action $a$ within the given context is estimated as the similarity between \(\mathcal{A}_a\) and \(\mathcal{X}_{t,a}\), as shown in (\ref{eq: e_t_a}). 
    \begin{equation}\label{eq: e_t_a}
        e_{t, a} = \delta(\mathcal{A}_{a}, \mathcal{X}_{t,a})
    \end{equation}
   	
    The action \(a_t\) is then selected according to the adopted exploration strategy, which integration will be explored in the following Subsections. The model receives a real-valued payoff \(r_{t,a_t}\), that is projected into the high-dimensional vector \(\mathcal{R}_{t,a_t}\) and used to update the HD-CB model as in (\ref{eq: update_A}). 
        \vspace{-0.5em}
    \begin{equation}
        \label{eq: update_A}
        \mathcal{A}_{a_t} = \mathcal{A}_{a_t} \oplus (\mathcal{X}_{t,a} \otimes \mathcal{R}_{t,a_t})
    \end{equation}
    Specifically, \(\mathcal{R}_{t,a_t}\) is generated using a thermometer encoding technique \cite{Rachkovskiy_2005} where the number of active elements increases with the magnitude of the scalar value.
   This encoding ensures that \(\mathcal{A}_{a_t}\) is adjusted closer to or further from the encoded context based on the received reward, learning at each iteration.
   In Figure \ref{fig:HD-CB_basics}, the collection of \(\mathcal{X}_{t,a}\) and   $e_{t,a}$ for all actions $a$ is denoted as \(\mathbf{\mathcal{X}_{t}}\) and $\mathbf{e_t}$, respectively. 
        
    Importantly, while this work employs real-valued HVs following the Multiply-Add-Permute-C implementation \cite{Kenny} from Gayler \cite{map}, the proposed methodology can be extended to other representations, such as complex or binary elements, offering flexibility to balance performance and computational complexity.

    
\subsection{HD-CB$_{\text{EPS}}$}\label{HDCB_eps}
    
Algorithm \ref{pseudocode: HDCUCB1} illustrates the $\text{HD-CB}_{\text{EPS}}$ model, which incorporates the approach outlined in the previous Section and employs the $\epsilon$-greedy exploration strategy. At each iteration, with a probability of \(1 - \epsilon\), the algorithm selects the action with the highest estimated reward \(e_{t,a}\), while with a probability of \(\epsilon\), it explores by selecting a random action. This simple approach requires tuning only one hyperparameter ($\epsilon$) and is well-suited for resource-constrained edge applications thanks to its low complexity.
However, as discussed in Section \ref{Background_and_Relatedworks}, the $\epsilon$-greedy strategy does not account for the confidence of the model in the expected payoffs of actions, potentially leading the algorithm to over-exploit actions with high uncertainty or under-explore potentially better alternatives.
For example, if an action has a high estimated reward but has been sampled only a few times, the algorithm may over-exploit it—mistakenly treating it as the optimal choice. In contrast, an action with a slightly lower but more reliably estimated reward could be unfairly neglected, leading to suboptimal long-term performance.
\begin{algorithm}[!t]
        \caption{$\text{HD-CB}_{\text{EPS}}$ Model}\label{pseudocode: HDCUCB1}
            \begin{algorithmic}[1]
                \STATE \textbf{Input:}\text{$\ \ \epsilon \in [0,1]$}
                \STATE \textbf{for} \text{$t = 1,2,..., T$}
                \STATE \textbf{\ \ \ for} \text{$a= 1,2,..., N$}
                \STATE   \text{\ \ \ \ \ \ Observe the context $\mathbf{x_{t,a}} \in \mathbb{R}^{d}$}
                \STATE   \text{\ \ \ \ \ \  Encoding the context in the $\mathcal{X}_{t,a}$ HV}
                \STATE   \text{\ \ \ \ \ \ \textbf{if} $t=1$: (Model Initialization) }
                \STATE   \text{\ \ \ \ \ \ \ \ \ $\mathcal{A}_{a} = \mathbf{0}_{D}$}
                \STATE   \text{\ \ \ \ \ \ \textbf{endif}}
                \STATE   \text{\ \ \ \ \ \  $e_{t, a} = \delta(\mathcal{A}_{a}, \mathcal{X}_{t})$}                
                \STATE   \text{\ \ \ \textbf{endfor}}
                \STATE   \text{\ \ \ $n = \text{uniform random in [0,1]}$ }
                \STATE   \text{\ \ \ \textbf{if} $n<\epsilon$: (Exploration) }      
                \STATE   \text{\ \ \ \ \ \ $a_{t}=\text{random  in the action space}$}
                \STATE   \text{\ \ \ \textbf{else}: \text{       \ \ \ \ (Exploitation)}}       
                \STATE   \text{\ \ \ \ \ \ $a_{t}=\argmax_{a} (e_{t, a})$}
                \STATE   \text{\ \ \ \textbf{endif} }       
                \STATE   \text{\ \ \ Observe a real-valued payoff $r_{t,a_t}$}
                 \STATE   \text{\ \ \ Encoding the reward in the $\mathcal{R}_{t,a_t}$ HV}
                \STATE   \text{\ \ \ $\mathcal{A}_{a_t}= \mathcal{A}_{a_{t}} \oplus ( \mathcal{X}_{t,a}\otimes\mathcal{R}_{t,a_t})$  }                
                \STATE \textbf{endfor}
            \end{algorithmic}
        \end{algorithm}

    
\subsection{HD-CB$_{\text{UNC}1}$}\label{HDCB_unc}

\begin{figure}[!t]
            \centering
            \includegraphics[width=1\columnwidth]{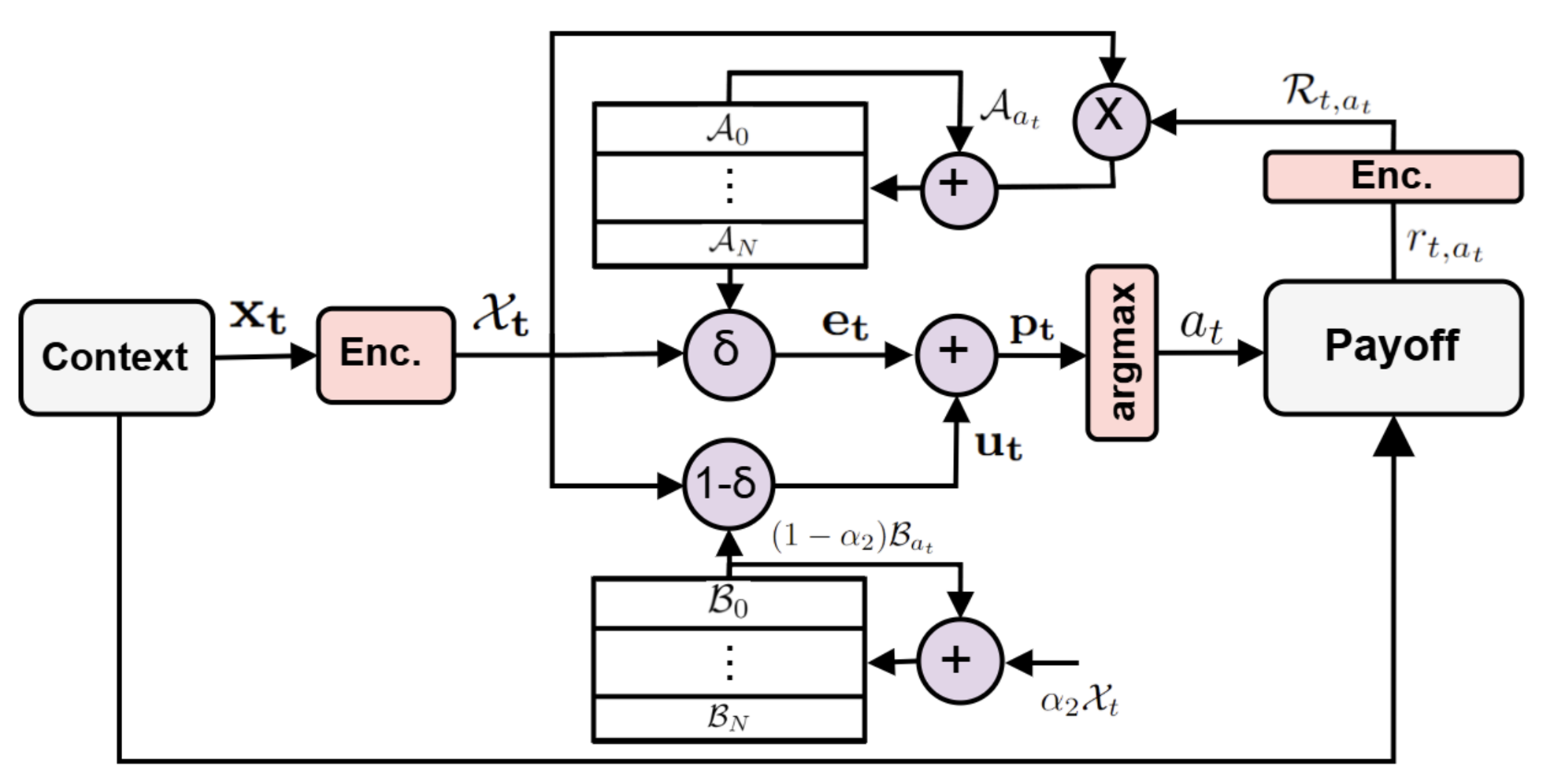}
             \caption{Schematic of HD-CB$_{\text{UNC1}}$. Each action \(a\) is associated with a confidence vector \(\mathcal{B}_a\). The vectors \(\mathbf{\mathcal{X}_{t}}\), \(\mathbf{e_t}\), \(\mathbf{u_t}\), and \(\mathbf{p_t}\) represent the context, estimated payoffs, uncertainties, and potential for all actions at time step \(t\).}
            \label{fig: HD-CB_UCB1}
        \end{figure}

    To address the limitations of the $\epsilon$-greedy strategy, this Section proposes the HD-CB$_{\text{UNC1}}$ model, depicted in Figure \ref{fig: HD-CB_UCB1} and detailed in Algorithm \ref{pseudocode: HD-CB_UCB1}. The core idea of this approach is to integrate a confidence measure into the expected payoff of each action. This is achieved by introducing an additional high-dimensional vector, \(\mathcal{B}_a\), for each action \(a\) that tracks the frequency of contexts in which the action is taken, regardless of the received reward.

    At each iteration, the model estimates the quality of actions using two key metrics: the estimated payoff (\(e_{t, a}\)) and the uncertainty (\(u_{t, a}\)). The estimated payoff is computed as previously shown in (\ref{eq: e_t_a}). 
    The uncertainty (\(u_{t, a}\)), on the other hand, is derived from the confidence (\(c_{t, a}\)), which is calculated as the similarity between \(\mathcal{B}_a\) and \(\mathcal{X}_{t,a}\), and quantifies how familiar the model is with the current context. 
    \begin{equation}\label{eq: c_t_a}
        c_{t, a} = \delta(\mathcal{B}_{a}, \mathcal{X}_{t,a})
    \end{equation}
    \begin{equation}\label{eq: u_t_a}
        u_{t, a} = 1 - c_{t, a}
    \end{equation}
    
    The sum of the estimated payoff and the uncertainty determines the potential expectation \((p_{t, a})\), as shown in (\ref{eq: p_t_a}). In this equation, the parameter \(\alpha\), referred to as the \textit{exploration factor}, regulates the balance between exploration and exploitation. Once \(p_{t, a}\) is calculated for all actions, the one with the highest potential value, \(a_t = \argmax_a(p_{t, a})\), is selected. 
    This formulation ensures that the $\text{HD-CB}_{\text{UNC1}}$ effectively combines the expected payoff and the confidence level in the decision-making process. 
    
    \begin{equation}
        \label{eq: p_t_a}
        p_{t, a} = e_{t, a} + \alpha\cdot u_{t, a} 
    \end{equation}

     After receiving the reward, the \(\mathcal{A}_{a_t}\) vector is updated following (\ref{eq: update_A}), while \(\mathcal{B}_{a_t}\) is adjusted using the exponential moving average in (\ref{eq: update_B}).
     Here, the \emph{smoothing factor} \((\alpha_2)\) serves as the second hyperparameter of the model and controls the weight of each new observation. A lower \(\alpha_2\) value results in more gradual updates, ensuring that the similarity between \(\mathcal{B}_{a_t}\) and the context vectors \(\mathcal{X}_{t,a}\) evolves slowly over time if the action is repeatedly chosen in similar contexts. This approach enables the model to adapt to evolving contexts by gradually incorporating new information while preserving past knowledge.
 
    \begin{equation}
        \label{eq: update_B}
        \mathcal{B}_{a_t} = (1 - \alpha_2) \mathcal{B}_{a_t} + \alpha_2 \mathcal{X}_{t,a}
    \end{equation}
        
        \begin{algorithm}[!t]
        \caption{HD-CB$_\text{unc1}$}\label{pseudocode: HD-CB_UCB1}
            \begin{algorithmic}[1]
                \STATE \textbf{Input:}\text{$\ \ \alpha \in \mathbb{R}_{+}$, $\alpha_2 \in \mathbb{R}_{+}$ }
                \STATE \textbf{for} \text{$t = 1,2,..., T$}
                \STATE \textbf{\ \ \ for} \text{$a= 1,2,..., N$}
                \STATE   \text{\ \ \ \ \ \ Observe the context $\mathbf{x_{t,a}} \in \mathbb{R}^{d}$}
                \STATE   \text{\ \ \ \ \ \  Encoding the context in the $\mathcal{X}_{t,a}$ HV}
                \STATE   \text{\ \ \ \ \ \ \textbf{if} $t=1$: (Model Initialization) }
                \STATE   \text{\ \ \ \ \ \ \ \ \ $\mathcal{A}_{a} = \mathbf{0}_{D}$}
                \STATE   \text{\ \ \ \ \ \ \ \ \ $\mathcal{B}_{a} = \mathbf{0}_{D}$}
                \STATE   \text{\ \ \ \ \ \ \textbf{endif}}
                \STATE   \text{\ \ \ \ \ \  $e_{t, a} = \delta(\mathcal{A}_{a}, \mathcal{X}_{t,a})$}

                \STATE   \text{\ \ \ \ \ \  $c_{t, a} = \delta(\mathcal{B}_{a}, \mathcal{X}_{t,a})$}
                \STATE   \text{\ \ \ \ \ \  $u_{t, a} = 1-c_{t,a}$}                
                
                \STATE   \text{\ \ \ \ \ \  $p_{t, a} = e_{t, a} + \alpha\cdot u_{t, a} $ }
                
                \STATE   \text{\ \ \ \textbf{endfor}}
                \STATE   \text{\ \ \ $a_{t}=\argmax_{a} (p_{t, a})$}
                \STATE   \text{\ \ \ Observe a real-valued payoff $r_{t,a_t}$}
                 \STATE   \text{\ \ \ Encoding the reward in the $\mathcal{R}_{t,a_t}$ HV}
                \STATE   \text{\ \ \ $\mathcal{A}_{a_t}= \mathcal{A}_{a_t} \oplus ( \mathcal{X}_{t,a}\otimes\mathcal{R}_{t,a_t})$  }
                \STATE   \text{\ \ \ $\mathcal{B}_{a_t} = (1 - \alpha_2) \mathcal{B}_{a_t} + \alpha_2 \mathcal{X}_{t,a}$ }
                
                \STATE \textbf{endfor}
            \end{algorithmic}
        \end{algorithm}

    
\subsection{HD-CB$_{\text{UNC2}}$: Controlling Model Updates with Confidence}\label{HDCB_unc2}

    In $\text{HD-CB}_{\text{UNC1}}$, the weight of each update to the \(\mathcal{B}_{a}\) HVs is controlled by the hyperparameter \(\alpha_2\), making the model adaptable to the specific characteristics and requirements of the problem at hand. The $\text{HD-CB}_{\text{UNC2}}$ algorithm eliminates this hyperparameter by introducing a dynamic update mechanism driven by the model's confidence. This mechanism leverages a \emph{thinning update} technique, where the confidence level determines the number of elements updated at each iteration. After selecting the action $a_t$, the confidence value $c_{t, a_t}$ is used to create a mask that randomly selects $c_{t, a_t} D$ elements of $\mathcal{B}_{a_t}$ to update, where $D$ represents the size of the HVs.
   
    This approach dynamically adjusts the weight of each update step based on the confidence of the model. When confidence is low, fewer elements are updated, allowing a slight increase in the similarity with the current context, as the model is less certain about the action-context relationship. Conversely, when confidence is high, more elements are updated, enabling the model to integrate new information more effectively. 
            
            
        
    This refined mechanism offers two significant advantages. First, it eliminates the need for the \(\alpha_2\) hyperparameter, simplifying the model and removing the requirement for expensive tuning procedures. Second, it does not involve operations between HVs beyond the basic HDC operators, allowing HVs to remain in the same representation space (e.g. binary). This characteristic is particularly advantageous for resource-constrained environments, as it makes the model easier to deploy with simpler hardware requirements while maintaining efficient and adaptive behaviour.


\subsection{$\text{HD-CB}_{\text{UNC3}}$: Reducing Memory Footprint by Permutation}\label{HDCB_unc3}
\begin{figure}[!t]
        \centering
        \includegraphics[width=1\columnwidth]{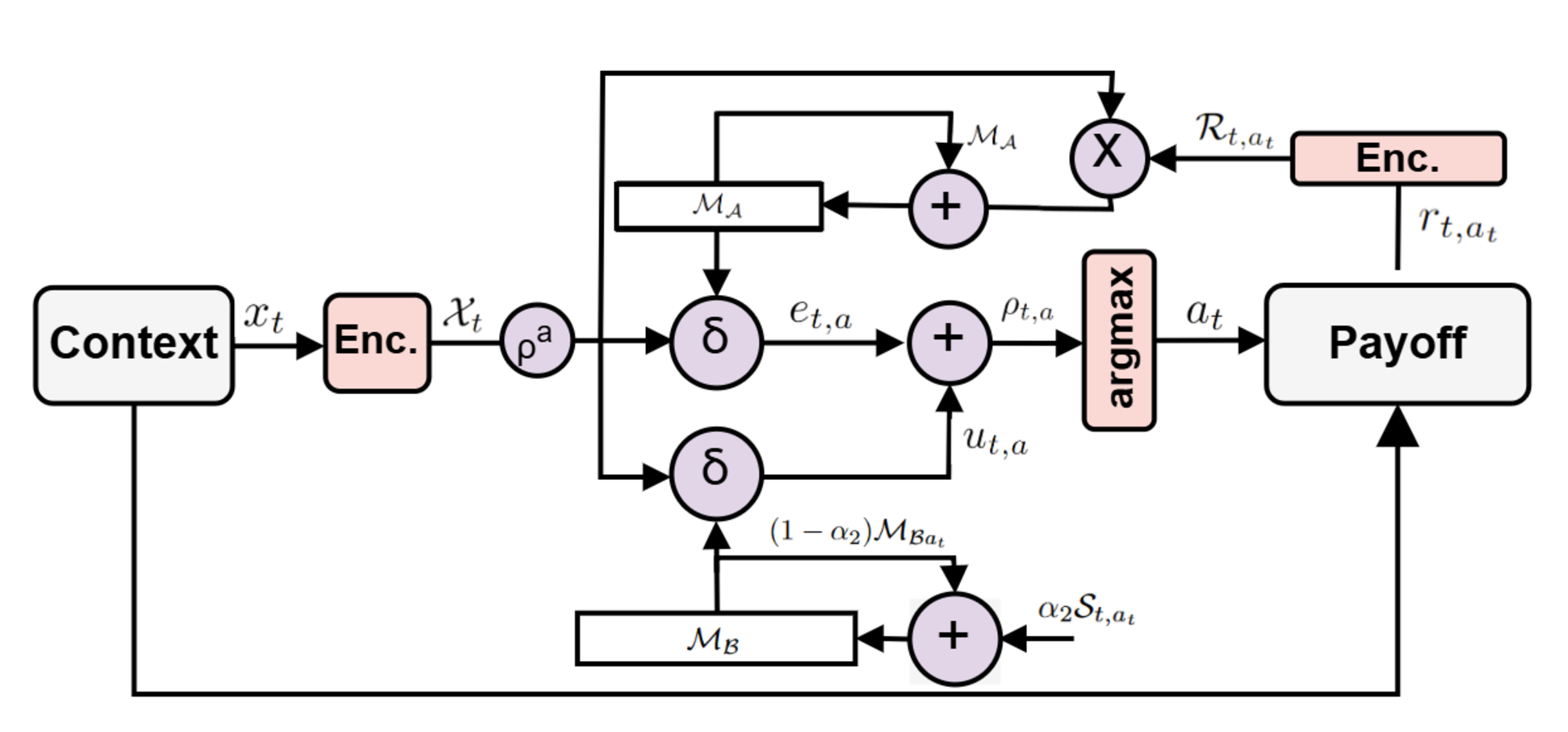}
       \caption{Schematic of HD-CB$_{\text{UNC3}}$. Memory overhead is reduced by using only two global HVs and leveraging the mathematical properties of permutation.}
        \label{fig: HD-CB3}
     \end{figure}

    The $\text{HD-CB}_{\text{UNC1}}$ and $\text{HD-CB}_{\text{UNC2}}$ models require two distinct HVs, \(\mathcal{A}_a\) and \(\mathcal{B}_a\), for each action \(a\) to effectively track the estimated reward and the uncertainty over time. However, as the number of actions and the size of the HVs increase, this can lead to significant memory overhead, potentially making these models impractical for resource-constrained embedded systems.
    To address this challenge, this Section proposes the $\text{HD-CB}_{\text{UNC3}}$ model, which optimizes memory usage by leveraging the mathematical properties of the permutation operator. Figure \ref{fig: HD-CB3} illustrates the schematic of HD-CB$_\text{UNC3}$.
    In this model, for a given action \(a\), a unique permutation \(\rho^a\) is applied to \(\mathcal{X}_{t,a}\), generating an HV \(\mathcal{S}_{t,a}\) that represents the context-action pair, as in (\ref{eq: state_action_HV}). 
    As discussed in Section \ref{Background_and_Relatedworks}, the permutation operation produces an orthogonal HV to the original one. This property ensures that context-action pairs remain distinguishable even when multiple pairs are superimposed, eliminating the need for dedicated HVs for each action.
    As a result, $\text{HD-CB}_{\text{UNC3}}$ requires only two global HVs: \(\mathcal{M}_\mathcal{A}\) and \(\mathcal{M}_\mathcal{B}\), initialized to zero. \
    \(\mathcal{M}_\mathcal{A}\) acts as a global repository for the outcomes of the selected context-action pairs, utilizing superposition and permutation operations to compactly encode this information, while \(\mathcal{M}_\mathcal{B}\) encodes the frequency of encountered context-action pairs, regardless of the rewards.
    \begin{equation}\label{eq: state_action_HV}
        \mathcal{S}_{t,a} = \rho^{a}(\mathcal{X}_{t,a})
    \end{equation}
    
    At each iteration, the potential value of each action \(p_{t,a}\) is computed as a combination of the estimated reward \(e_{t,a}\) and the uncertainty \(u_{t,a}\), calculated as shown in (\ref{eq: e_t_a_2}) and (\ref{eq: u_t_a_2}), respectively.
    \begin{equation}
        \label{eq: e_t_a_2}
        e_{t, a} = \delta(\mathcal{M}_{\mathcal{A}_{a}}, \mathcal{S}_{t,a})
    \end{equation}
    \begin{equation}
        \label{eq: u_t_a_2}
        u_{t, a} = 1 - \delta(\mathcal{M}_{\mathcal{B}_{a}}, \mathcal{S}_{t,a})
    \end{equation}
    
   After receiving the reward, the model updates both \(\mathcal{M}_\mathcal{A}\) and \(\mathcal{M}_\mathcal{B}\) as described in (\ref{eq: update_A_2}) and (\ref{eq: update_B_2}).

    \begin{equation}
        \label{eq: update_A_2}
        \mathcal{M}_{\mathcal{A}_{a_t}} = \mathcal{M}_{\mathcal{A}_{a_t}} \oplus (\mathcal{S}_{t,a_t} \otimes \mathcal{R}_{t,a_t})
    \end{equation}
    \begin{equation}
        \label{eq: update_B_2}
        \mathcal{M}_{\mathcal{B}_{a_t}} = (1 - \alpha_2) \mathcal{M}_{\mathcal{B}_{a_t}} + \alpha_2 \mathcal{S}_{t,a_t}
    \end{equation}
    Notably, the thinning update from $\text{HD-CB}_{	\text{UNC2}}$ can also be applied here, eliminating scalar multiplications and further simplifying the update process for embedded applications.

        
    \section{Results} \label{Results}
    To evaluate the performance of the presented HD-CB models, we conducted a series of experiments across various datasets and evaluation frameworks, focusing on key performance metrics.
    First, we assessed the online learning capabilities of the models on synthetic datasets by analyzing average reward, cumulative regret, and convergence time, then we benchmarked the HD-CB models against state-of-the-art approaches using the Open Bandit Pipeline (OBP) framework \cite{zr_obp} for off-policy evaluation and the MovieLens-100k dataset for tests in real-world recommendation scenarios. Finally, we analyzed the effect of the HVs size on average reward and execution time, highlighting the trade-offs between computational efficiency and model accuracy. The computational complexity of HD-CB was also compared with traditional algorithms, showcasing its advantages in resource-constrained environments.
    
     \begin{figure*}[!t]
        \centering
        \includegraphics[width=1.7\columnwidth]{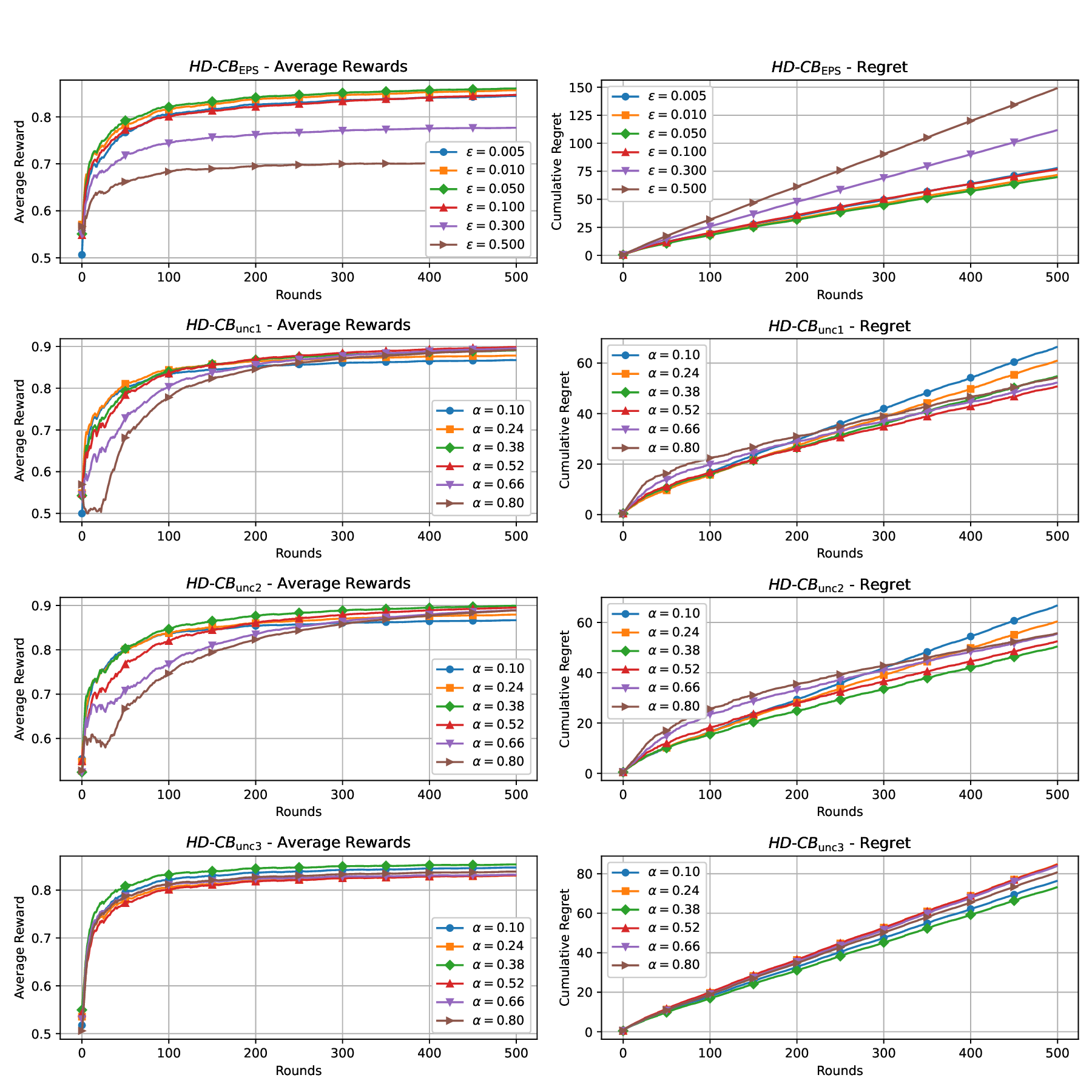}
        \caption{Average reward and cumulative regret trends for HD-CB models as a function of the exploration parameters ($\epsilon$ for $\text{HD-CB}_{\text{EPS}}$ and $\alpha$ for the uncertainty-driven models), over 500 iterations, averaged across 100 synthetic datasets. The experiments were conducted with a hypervector size of 1000, 20 actions, continuous rewards in the range [0,1], and a context feature vector size of 10. 
        }
        \label{fig: rewards}
    \end{figure*}

    \subsection{Online Learning Capabilities}
    To assess the online learning capabilities of the HD-CB models, we designed synthetic datasets to emulate CB problems. In these scenarios, the models iteratively observe new contexts, select actions, and receive rewards, mimicking real-world decision-making processes. These controlled environments offer a robust and replicable framework for evaluating the models, enabling a detailed analysis of their adaptability and learning performance under consistent and well-defined conditions.

    In this analysis, we evaluated the average reward and cumulative regret over time as key metrics. 
    The average reward measures how effectively the model selects actions that maximize total payoff, while cumulative regret quantifies the difference between the rewards of optimal actions and those chosen by the model. A lower cumulative regret reflects the capability of the model to adapt quickly and align its decision-making with the optimal strategy over time, demonstrating efficient learning and exploration.

    \begin{table*}[!t]
        \caption{Off-policy evaluation for HD-CB models and traditional CB algorithms on synthetic datasets with binary rewards. The experiments were conducted with varying numbers of actions (\(N = 10, 15, 20\)) and context feature vector sizes (\(d = 5, 10\)). 
        Average rewards \(\pm\) standard deviation) were calculated for each configuration across 50 datasets generated using OBP.}        
        \label{tab:binary_rewards}
        \centering
        \resizebox{1.8\columnwidth}{!}{\begin{tabular}{c c | c c c c c c}
        \textbf{$\mathbf{N}$} & \textbf{$\mathbf{d}$} & \textbf{LinEPS} & \textbf{LinUCB} & \textbf{HD-CB$_{\epsilon}$} & \textbf{HD-CB$_{\text{UNC}1}$} & \textbf{HD-CB$_{\text{UNC}2}$} & \textbf{HD-CB$_{\text{UNC}3}$} \\ 
        \midrule
        10 & 5  & 0.607 ± 0.067 & 0.674 ± 0.061 & 0.670 ± 0.059 & 0.691 ± 0.062 & \textbf{0.691 ± 0.065} & 0.679 ± 0.062 \\ 
        10 & 10 & 0.623 ± 0.087 & 0.704 ± 0.055 & 0.690 ± 0.067 & 0.707 ± 0.068 & \textbf{0.711 ± 0.072} & 0.690 ± 0.068 \\ 
        15 & 5  & 0.700 ± 0.065 & 0.753 ± 0.050 & 0.758 ± 0.041 & 0.787 ± 0.046 & \textbf{0.794 ± 0.044} & 0.772 ± 0.046 \\ 
        15 & 10 & 0.619 ± 0.090 & 0.695 ± 0.074 & 0.696 ± 0.075 & 0.718 ± 0.078 & \textbf{0.724 ± 0.078} & 0.696 ± 0.078 \\ 
        20 & 5  & 0.680 ± 0.067 & 0.715 ± 0.059 & 0.722 ± 0.062 & 0.760 ± 0.062 & \textbf{0.761 ± 0.063} & 0.737 ± 0.060 \\ 
        20 & 10 & 0.564 ± 0.098 & 0.648 ± 0.088 & 0.647 ± 0.092 & 0.668 ± 0.093 & \textbf{0.674 ± 0.091} & 0.645 ± 0.090 \\ \bottomrule
        \end{tabular}}
    \end{table*}
    
    Figure \ref{fig: rewards} illustrates the average performance of each algorithm across 100 synthetic datasets, configured with a hypervector size of 1000, 20 actions, continuous reward in [0,1], a context feature vector size of 10, and varying the hyperparameters $\epsilon$ and $\alpha$. The results are presented over 500 iterations. All the subplots consistently demonstrate the ability of the model to learn and converge towards optimal solutions, as reflected by the average reward approaching 1.

    Similar to the traditional CB algorithms, the performance of all HD-CB models is significantly influenced by their respective exploration parameters (i.e., \(\epsilon\) in $\text{HD-CB}_{\text{EPS}}$ and \(\alpha\) in the uncertainty-driven variants). Small parameter values (e.g., \(\epsilon = 0.01\) or \(\alpha = 0.1\)) result in insufficient exploration, leading to suboptimal performance characterized by slow reward growth and minimal reduction in cumulative regret. Conversely, high parameter values lead to over-exploration, where the models continue to explore unnecessarily even after identifying the optimal action, thereby reducing the overall cumulative reward. Notably, $\text{HD-CB}_{\text{EPS}}$ exhibits the highest sensitivity to its exploration parameter, requiring careful tuning for optimal performance.

    When comparing the models in terms of average rewards, $\text{HD-CB}_{\text{UNC2}}$ demonstrates the highest performance, achieving an average reward of approximately $0.908$ and a cumulative regret of $50.36$ with \(\alpha = 0.38\). The $\text{HD-CB}_{\text{UNC1}}$ and $\text{HD-CB}_{\text{UNC3}}$ models follow with maximum average rewards of $0.899$ and $0.863$ and cumulative regrets of $50.70$ and $69.13$ when using $\alpha$ equal to $0.52$ and $0.38$, respectively. Finally, $\text{HD-CB}_{\text{EPS}}$ achieves a maximum reward of $0.85$ and a regret of $69.76$ when $\epsilon=0.05$. This ranking aligns with the computational complexity of the models: $\text{HD-CB}_{\text{EPS}}$ and $\text{HD-CB}_{\text{UNC3}}$re the simplest and least computationally demanding, whereas $\text{HD-CB}_{\text{UNC1}}$ and $\text{HD-CB}_{\text{UNC2}}$ are more complex and deliver higher performance.
    
    Overall, these results demonstrate the online learning capabilities of HD-CB models and underscore the effectiveness of uncertainty-driven exploration strategies in maximizing cumulative performance over time.

    \subsection{Off-Policy Evaluation}\label{OPB}

    To evaluate the performance of the proposed HD-CB models against traditional CB algorithms illustrated in Section \ref{Background_and_Relatedworks}, we utilized the Open Bandit Pipeline (OBP) \cite{zr_obp}: an open-source framework designed to generate synthetic datasets and provide standardized off-policy evaluations of bandit algorithms. Off-policy evaluation (OPE) plays a critical role in reinforcement learning and contextual bandit settings, offering a safe, efficient, and reproducible way to estimate the performance of a new policy.
    In an online setting, in fact, only the reward for the selected action is observed at each time step \cite{lattimore2020bandit}, making offline evaluation on logged data essential for obtaining unbiased performance estimations of target policies. OPE is particularly important in high-stakes domains, such as healthcare, where deploying an untested model could lead to costly or risky outcomes.
    Using the OBP framework, we conducted tests on 50 different synthetic datasets with binary rewards, varying the number of actions ($N$) and the context feature vector size ($d$). Table \ref{tab:binary_rewards} summarizes the average rewards achieved by all HD-CB models compared to LinearEPS and LinUCB algorithms. For all the algorithms, we performed a grid search over the $\epsilon$, $\alpha$ and $\alpha_2$ (only for HD-CB$_{\text{UNC1}}$ and HD-CB$_{\text{UNC3}}$) parameters in the range \([0.1, 0.2, \ldots, 0.9, 1]\). Additionally, we searched for the optimal number of quantization levels from the set \([10, 15, 20]\) for the HD-CB models.
    
    When employing an $\epsilon$-greedy strategy, HD-CB$_{\epsilon}$ consistently outperforms LinEPS, with an average improvement of 9\% across all dataset configurations. This result is particularly significant as it directly compares the simplest versions of both frameworks, utilizing the same exploration strategy. The superior performance of the baseline HD-CB framework underscores the potential of HDC over traditional linear algorithms, even when paired with simple exploration strategies.
    
    \begin{figure}[!t]
        \centering
        \includegraphics[width=0.7\columnwidth]{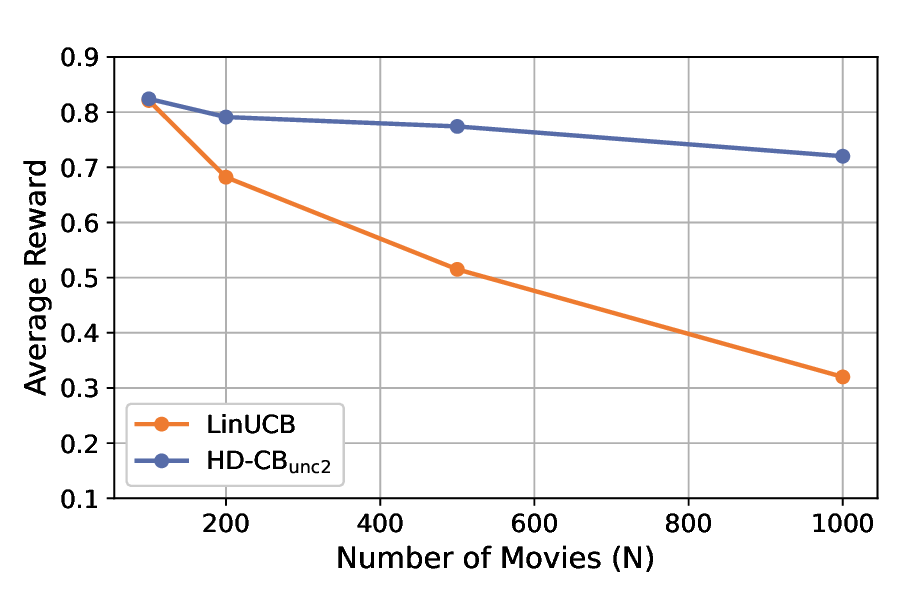}
        \caption{Results obtained on the MovieLens-100k dataset by the traditional LinUCB algorithm and the proposed HD-CB2 model, varying the number of movies.}
        \label{fig:movielens_results}
    \end{figure}
    The uncertainty-driven HD-CB models, including HD-CB$_{\text{UNC1}}$, HD-CB$_{\text{UNC2}}$, and HD-CB$_{\text{UNC3}}$, consistently demonstrate comparable or superior performance to LinUCB across all tested configurations. Notably, HD-CB$_{\text{UNC2}}$ achieves the highest average reward across all configurations, outperforming LinUCB by an average of 4\% thanks to its confidence-driven update mechanism. 
    HD-CB$_{\text{UNC3}}$, which leverages a more compact representation through permutation-based superposition, also shows competitive results while using significantly less memory. This makes it an attractive choice for resource-constrained environments where memory efficiency is critical.
    
    Overall, these findings validate the effectiveness of the proposed HD-CB framework for modelling and solving CB problems, demonstrating comparable or superior performance to traditional algorithms. By mapping the problem into a high-dimensional space, HD-CB effectively separates context-action pairs, enhancing learning efficiency and facilitating more accurate decision-making.

    \subsection{Evaluation on the MovieLens-100k Dataset}
    To evaluate the performance of HD-CB models on real-world data, we conducted tests using the MovieLens-100k dataset. MovieLens is a widely used benchmark for recommendation systems that consists of 100,000 ratings from 943 users for 1,682 movies, with ratings ranging from 1 (lowest) to 5 (highest). We considered ratings of 4 or higher as positive, allowing us to frame the recommendation task as a CB problem with binary rewards. Specifically, when a user gives a rating of 4 or higher, the model receives a reward of 1; otherwise, the reward is set to 0.
    Using a real dataset such as MovieLens is ideal for validating the performance of our model in a real-world context. 

    Figure \ref{fig:movielens_results} shows the average rewards achieved by the traditional LinUCB algorithm and our best-performing HD-CB$_{unc2}$ model, averaged over 50 runs with the number of possible actions (i.e., movies available for recommendation) ranging from 100 to 1000. Like previous experiments, a grid search determined the optimal values for the $\alpha$ and $\alpha_2$ hyperparameters.
    The results indicate that the HD-CB model consistently outperforms LinUCB, particularly as the number of actions increases. Both models achieve similar average rewards for \(N = 100\), with HD-CB$_{\text{unc2}}$ slightly outperforming Linear UCB (0.824 vs. 0.821). However, as \(N\) increases to 200, 500, and 1000, the gap between the two models becomes more pronounced. For \(N = 1000\), the average reward of the HD-CB$_{\text{unc2}}$ model (0.720) is significantly higher than that of LinUCB (0.320), demonstrating the robustness of the HD-CB$_{\text{unc2}}$ model in handling larger action spaces and its potential for real-world recommendation scenarios.

    \subsection{Convergence Time Analysis}
    \begin{figure}[!t]
        \centering
        \includegraphics[width=1\columnwidth]{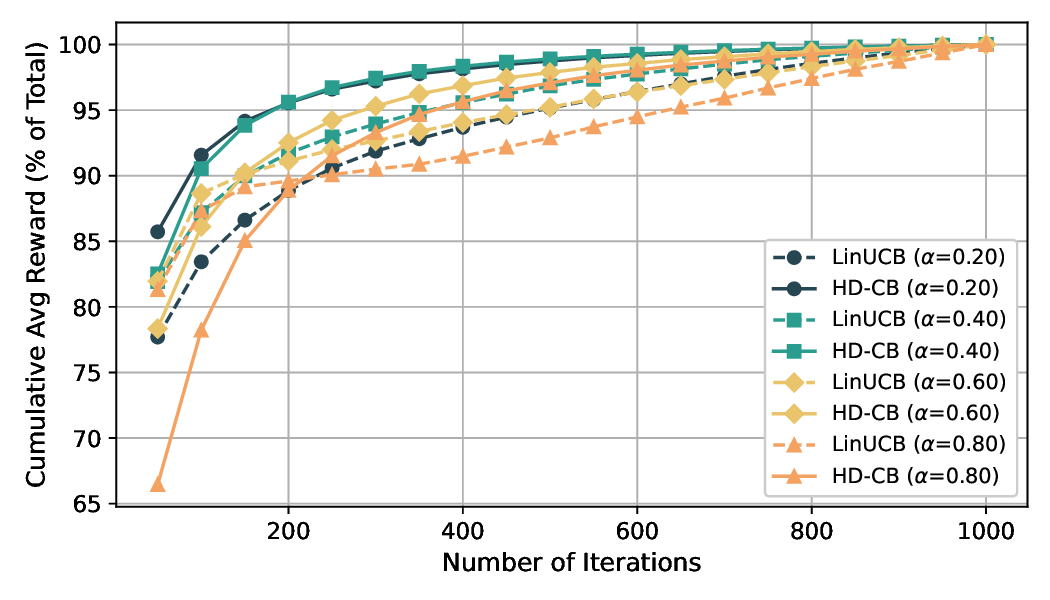}
        \caption{Convergence time as a function of \(\alpha\), comparing the LinUCB algorithms and the HD-CB$_{\text{UNC1}}$ model.}
        \label{fig: convergence_time}
    \end{figure}
    In online decision-making problems, convergence time is a critical aspect of algorithm performance. Unlike traditional offline learning, where models are trained before deployment, online learning algorithms must adapt and improve in real time, using rewards collected at each iteration to make better future decisions. Therefore, a faster convergence time translates to improved decision quality early in the learning process, which is particularly important in applications where rapid adaptation is essential, such as personalized recommendations, healthcare, or autonomous systems.
    
    HDC has demonstrated faster convergence times and reduced data requirements in many learning tasks \cite{faster1, faster2}. This advantage stems from the HDC abilities to project problems into high-dimensional space, where objects can be more easily separated, making the learning process more efficient. Motivated by these properties, in this Subsection, we investigate the convergence performance of the proposed HD-CB algorithms and compare them with traditional LinUCB models.

    We conducted experiments using synthetic datasets, averaging the results over 100 different runs while varying the exploration parameter \(\alpha\). Figure \ref{fig: convergence_time} presents the results in terms of cumulative reward as a percentage of the final reward over time, providing a clear visualization of each model's convergence rate towards near-optimal decision-making. 
    Overall, the HD-CB models consistently converge faster than their LinUCB counterparts, regardless of the value of \(\alpha\). This highlights the efficiency and adaptability of HD-CB in online decision-making environments, making it a promising approach for scenarios where rapid convergence is essential.

    \subsection{Computational Complexity Analysis}\label{Complexity_Analysis}
    
    The computational complexity of CB algorithms can be critical in resource-constrained applications requiring rapid and efficient decision-making. Traditional algorithms rely on the inversion of $N$ matrices of dimensions $d \times d$ at each time step, where $N$ is the number of actions, and $d$ is the size of the context feature vector. This results in an asymptotic complexity of $\mathcal{O}(d^3)$ for each matrix inversion, imposing a significant computational burden that can create performance bottlenecks and make these algorithms less suitable for time-sensitive applications.

    This complexity can be reduced to $\mathcal{O}(d^2)$ by employing the incremental update mechanism proposed in \cite{LINUCBOPT}, which leverages the Sherman-Morrison theorem to efficiently update the inverse matrices without recomputing them from scratch. 
    
    The HD-CB framework eliminates the need for matrix inversion entirely. By representing each action with a dedicated hypervector and using simple arithmetic operations in the high-dimensional space, HD-CB reduces the complexity with respect to the context size to $\mathcal{O}(d)$.  Additionally, HD-CB maintains linear complexity in the number of actions ($\mathcal{O}(N)$) and introduces a dependence on HV size ($\mathcal{O}(D)$), making its computational complexity primarily determined by $D$. 
    The linear dependence of HD-CB on $d$ provides a substantial advantage. Traditional CB algorithms often require dimensionality reduction techniques to avoid the computational burden of high-dimensional contexts, especially in domains such as web applications or personalized recommendations, where contexts may represent detailed user profiles. HD-CB removes this problem, simplifying the pipeline and avoiding additional pre-processing methods, making it highly suitable for practical real-time and resource-constrained applications.

    Table \ref{tab:complexity} summarizes the computational complexity associated with the three algorithms, highlighting their respective dependencies on context size, number of actions, and HV size.

    \begin{table}[h!]
    \caption{Computational Complexity of Contextual Bandit Algorithms}
    \label{tab:complexity}
    \centering
    \begin{tabular}{l|ccc}
    \textbf{Algorithm} & \textbf{Context Size ($d$)} & \textbf{N. Actions ($N$)} & \textbf{HV Size ($D$)} \\ 
    \midrule
    Linear CB       & $\mathcal{O}(d^3)$ & $\mathcal{O}(N)$ & -- \\ 
    Opt. Linear CB        & $\mathcal{O}(d^2)$ & $\mathcal{O}(N)$ & -- \\
    HD-CB                     & $\mathcal{O}(d)$ & $\mathcal{O}(N)$ & $\mathcal{O}(D)$ \\ \bottomrule
    \end{tabular}
    \end{table}
  
    \begin{figure*}[!t]
        \centering
        \includegraphics[width=1.6\columnwidth]{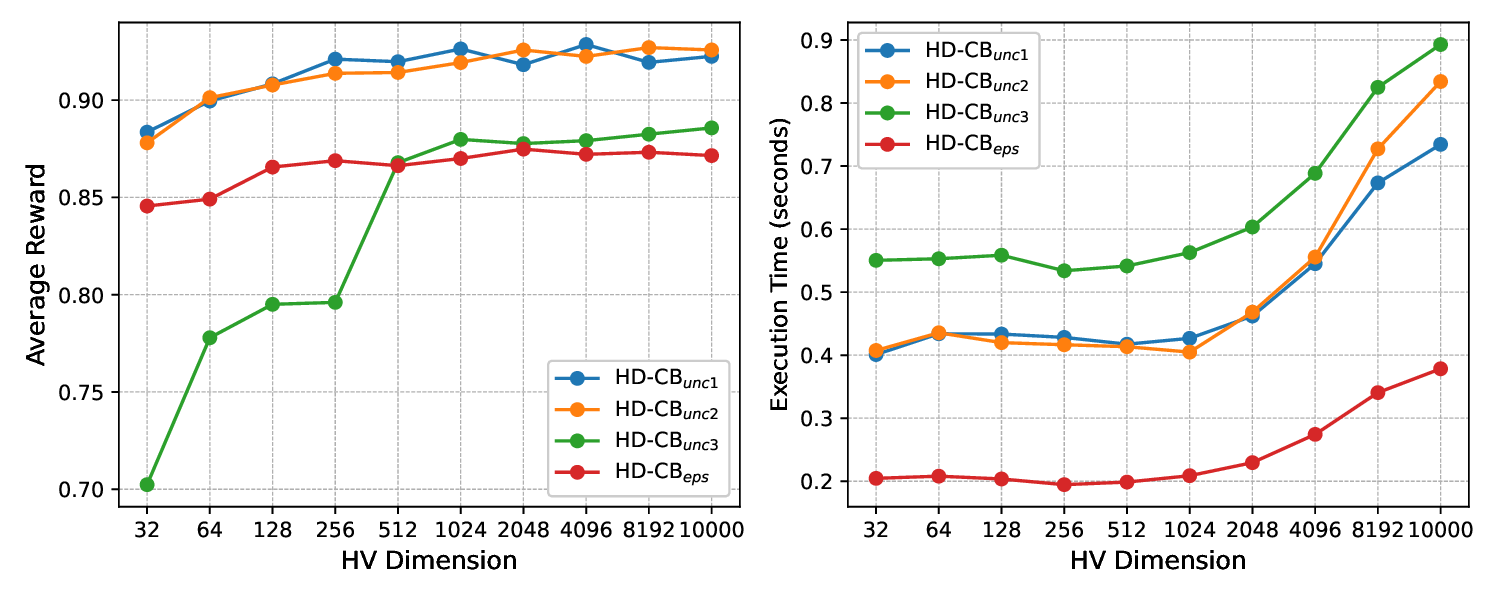}
        \caption{Impact of HV size on average reward and execution time for the proposed HD-CB models. The figure highlights the trade-offs between accuracy and computational efficiency as HV size varies from 32 to 10,000.}

        \label{fig: hdcb_sizes}
    \end{figure*}
    \subsection{Scalability Analysis}       
    The HV size plays a crucial role in determining the performance and efficiency of HDC models. Larger HV sizes enhance the representational capacity of the high-dimensional space, allowing the model to encode and discriminate between distinct objects more effectively. This increased capacity can significantly improve accuracy, particularly in scenarios with high variability or complex datasets \cite{schlegel2022comparison, neubert2019introduction}. However, larger HV sizes also lead to increased computational complexity and memory overhead, making it essential to balance these factors when designing and deploying HDC-based systems.
    
    To assess the impact of HV size on the performance of the proposed HD-CB models, we conducted experiments on 100 synthetic datasets, with the context feature vector size and the number of actions both set to 10. The HV size was varied from 32 to 10,000, and the optimal values for the hyperparameters $\epsilon$, $\alpha$, and $\alpha_2$ were determined through grid search, as for previous experiments. Figure \ref{fig: hdcb_sizes} illustrates the average reward and execution time trends as functions of HV size for the different HD-CB models. 

     From a performance perspective, the HD-CB$_{\text{EPS}}$, HD-CB$_{\text{UNC1}}$, and HD-CB$_{\text{UNC2}}$ models demonstrates notable robustness to reductions in HV dimensionality. Their performance decreases by only $0.41\%$, $1.61\%$, and $1.40\%$, respectively, when the HV size is reduced from 10.000 to 256. As the HV size is further decreased to 64, the average reward dropped by $4\%$, $5\%$, and $6\%$, respectively, and by $7\%$, $8\%$, and $8.6\%$ when the HV size reached 32.
     This stability is likely due to the independent representation of each action using distinct HVs, which maintains discrimination capabilities even at lower dimensionalities and ensures the models can deliver high accuracy even in edge computing real-time decision-making systems.
    
     On the other hand, HD-CB$_{\text{UNC3}}$ exhibits the highest sensitivity to reductions in HV size, with its average reward decreasing by $10\%$ when the HV size is reduced from 10,000 to 256 and by $16\%$ when further decreased to 64. This sensitivity is expected, as HD-CB$_{\text{UNC3}}$ uses only two HVs for the entire model and relies on permutation to encode actions. The shared HVs are updated incrementally for all context-action pairs, which quickly saturates the capacity of the high-dimensional space at lower dimensions. As a result, larger HV sizes are essential to maintain the ability to encode and discriminate between different actions effectively.
     Despite this sensitivity, HD-CB$_{\text{UNC3}}$ reduces memory overhead by a factor proportional to $(2N)$ compared to HD-CB$_{\text{UNC1}}$ and HD-CB$_{\text{UNC2}}$, resulting a compelling choice in memory-constrained scenarios.
    
     In terms of execution time, HD-CB${\text{EPS}}$ is the fastest, owing to its simplicity and the use of a single hypervector per action. It is followed by HD-CB${\text{UNC1}}$ and HD-CB${\text{UNC2}}$, whose execution times are approximately double due to the additional overhead of maintaining and updating two hypervectors per action. Moreover, HD-CB$_{\text{UNC2}}$ exhibited slightly higher runtime compared to HD-CB$_{\text{UNC1}}$ due to the additional computational steps involved in the thinning update.
     HD-CB$_{\text{UNC3}}$, on the other hand, exhibited the slowest execution time among all models. This is primarily because it requires an additional $N$ permutations at each iteration to encode actions using a reduced number of shared hypervectors. 
    
     Overall, the scalability analysis emphasizes the trade-offs between performance and execution time across the different HD-CB models, underscoring the importance of selecting the optimal model based on specific application requirements. For scenarios that demand computational efficiency and rapid decision-making, such as real-time systems, HD-CB$_{\text{EPS}}$ and HD-CB$_{\text{UNC1}}$ emerge as the most suitable choices due to their robust performance and lower computational demands. On the other hand, HD-CB$_{\text{UNC3}}$ stands out in memory-constrained environments, offering a compact solution with significant memory savings while maintaining acceptable accuracy. This flexibility to adapt to diverse operational constraints highlights the practical utility and versatility of the HD-CB framework in addressing real-world CB problems effectively and efficiently.

            
\section{Conclusions}\label{Conclusions}

    In this paper, we introduced the Hyperdimensional Contextual Bandits (HD-CB) framework, the first exploration of Hyperdimensional Computing in the literature for modelling and solving sequential decision-making problems in the Contextual Bandits domain. The HD-CB framework leverages the mathematical properties of high-dimensional spaces to encode the state of the environment and model each action using dedicated hypervectors. At each iteration, these hypervectors are used to identify the action with the highest expected reward, with dynamic updates based on the received payoff ensuring adaptability over time.
    In this way, HD-CB replaces computationally expensive ridge regression procedures in traditional CB algorithms with simple, highly parallel vector operations in the high-dimensional space, making it well-suited for deployment in resource-constrained systems and as a perfect candidate for hardware acceleration.
    
    We detailed HD-CB in four distinct variants, each addressing specific challenges and requirements in CB problems, with the flexibility to select the most suitable implementation based on application needs and available system resources.
    HD-CB\(_{\text{EPS}}\) integrates a simple and lightweight \(\epsilon\)-greedy exploration strategy, making it suitable for resource-constrained systems and less complex environments. HD-CB\(_{\text{UNC1}}\) improves decision accuracy by incorporating a confidence-driven exploration mechanism but at the cost of added complexity and the introduction of a second hyperparameter, \(\alpha_2\). HD-CB\(_{\text{UNC2}}\) simplifies this approach by dynamically controlling updates based on model confidence, eliminating the need for \(\alpha_2\). Finally, HD-CB\(_{\text{UNC3}}\) addresses scenarios where memory overhead is critical, achieving substantial memory reductions while maintaining lower but competitive performance. 

    Extensive testing validated the HD-CB framework across synthetic and real-world datasets, demonstrating its effectiveness in terms of online learning capabilities, average reward, cumulative regret, and convergence speed. Comparisons with traditional state-of-the-art algorithms revealed that HD-CB models consistently achieve competitive performance, faster convergence times, and lower computational complexity. 
    Finally, we analyzed the scalability of the models by varying hypervector sizes and measuring the impact on reward and execution time, providing practical guidelines for balancing accuracy, computational efficiency, and memory requirements based on specific application needs.
     
    Collectively, these contributions represent a significant advancement in the field, offering an entirely new framework for integrating HDC into CB problems. By introducing both a novel baseline model and a set of versatile variants, this work lays the groundwork for future research and practical applications.     
    Future research could explore dynamic tuning of hyperparameters, the integration of advanced exploration strategies, and the deployment of HD-CB models in new real-world domains requiring efficient and scalable online learning capabilities.  Moreover, developing a hybrid HD-CB approach, where shared hypervectors capture shared relationships across actions, drawing inspiration from traditional Hybrid LinUCB models, could further enhance the efficiency and effectiveness of the framework.
    Finally, exploring hardware acceleration for HD-CB models presents a promising direction, leveraging their inherent parallelism to reduce computational overhead drastically. This advancement could open up new opportunities for CB problems in resource-constrained environments and real-time applications, where achieving low latency and high energy efficiency is paramount.

    \bibliographystyle{ieeetr}
    \bibliography{bibliography.bib}

\begin{thebibliography}{10}

\bibitem{CB_review}
D.~Bouneffouf, I.~Rish, and C.~Aggarwal, ``Survey on applications of multi-armed and contextual bandits,'' in {\em 2020 IEEE Congress on Evolutionary Computation (CEC)}, p.~1–8, IEEE Press, 2020.

\bibitem{li2010contextual}
L.~Li, W.~Chu, J.~Langford, and R.~E. Schapire, ``A contextual-bandit approach to personalized news article recommendation,'' in {\em Proceedings of the 19th international conference on World wide web}, pp.~661--670, 2010.

\bibitem{Netflix}
F.~Amat, A.~Chandrashekar, T.~Jebara, and J.~Basilico, ``Artwork personalization at netflix,'' in {\em Proceedings of the 12th ACM Conference on Recommender Systems}, RecSys '18, (New York, NY, USA), p.~487–488, Association for Computing Machinery, 2018.

\bibitem{MICROSOFT}
A.~Agarwal, S.~Bird, M.~Cozowicz, M.~Dud{\i}k, L.~Hoang, J.~Langford, L.~Li, D.~Melamed, G.~Oshri, S.~Sen, {\em et~al.}, ``Multiworld testing decision service: A system for experimentation, learning, and decision-making,'' {\em Whitepaper of Microsoft}, pp.~1--40, 2016.

\bibitem{NYT}
T.~N.~Y. Times, ``How the new york times is experimenting with recommendation algorithms,'' 2019.
\newblock Accessed: 2024-08-26.

\bibitem{GAME}
N3twork, ``How we boosted app revenue by 10\% with real-time personalization,'' 2020.
\newblock Accessed: 2024-08-26.

\bibitem{angioli_automatic}
M.~Angioli, M.~Barbirotta, A.~Mastrandrea, S.~Jamili, and M.~Olivieri, ``Automatic hardware accelerators reconfiguration through linearucb algorithms on a risc-v processor,'' in {\em 2023 18th Conference on Ph.D Research in Microelectronics and Electronics (PRIME)}, pp.~169--172, 2023.

\bibitem{EDGE1}
M.~V. Ngo, T.~Luo, H.~Chaouchi, and T.~Q. Quek, ``Contextual-bandit anomaly detection for iot data in distributed hierarchical edge computing,'' in {\em 2020 IEEE 40th International Conference on Distributed Computing Systems (ICDCS)}, pp.~1227--1230, IEEE, 2020.

\bibitem{EDGE2}
C.-S. Yang, R.~Pedarsani, and A.~S. Avestimehr, ``Edge computing in the dark: Leveraging contextual-combinatorial bandit and coded computing,'' {\em IEEE/ACM Transactions on Networking}, vol.~29, no.~3, pp.~1022--1031, 2021.

\bibitem{MD1}
A.~Durand, C.~Achilleos, D.~Iacovides, K.~Strati, G.~D. Mitsis, and J.~Pineau, ``Contextual bandits for adapting treatment in a mouse model of de novo carcinogenesis,'' in {\em Machine learning for healthcare conference}, pp.~67--82, PMLR, 2018.

\bibitem{MD2}
H.~Bastani and M.~Bayati, ``Online decision making with high-dimensional covariates,'' {\em Operations Research}, vol.~68, no.~1, pp.~276--294, 2020.

\bibitem{Kenny}
K.~Schlegel, P.~Neubert, and P.~Protzel, ``A comparison of vector symbolic architectures,'' {\em Artificial Intelligence Review}, vol.~55, no.~6, pp.~4523--4555, 2022.

\bibitem{RW_review_1}
C.-Y. Chang, Y.-C. Chuang, C.-T. Huang, and A.-Y. Wu, ``Recent progress and development of hyperdimensional computing (hdc) for edge intelligence,'' {\em IEEE Journal on Emerging and Selected Topics in Circuits and Systems}, 2023.

\bibitem{RW_review_2}
D.~Kleyko, M.~Davies, E.~P. Frady, P.~Kanerva, S.~J. Kent, B.~A. Olshausen, E.~Osipov, J.~M. Rabaey, D.~A. Rachkovskij, A.~Rahimi, {\em et~al.}, ``Vector symbolic architectures as a computing framework for emerging hardware,'' {\em Proceedings of the IEEE}, vol.~110, no.~10, pp.~1538--1571, 2022.

\bibitem{RW_review_3}
D.~Kleyko, D.~Rachkovskij, E.~Osipov, and A.~Rahimi, ``A survey on hyperdimensional computing aka vector symbolic architectures, part ii: Applications, cognitive models, and challenges,'' {\em ACM Computing Surveys}, vol.~55, no.~9, pp.~1--52, 2023.

\bibitem{li2011unbiased}
L.~Li, W.~Chu, J.~Langford, and X.~Wang, ``Unbiased offline evaluation of contextual-bandit-based news article recommendation algorithms,'' in {\em Proceedings of the fourth ACM international conference on Web search and data mining}, pp.~297--306, 2011.

\bibitem{zr_obp}
{ZOZO Research}, ``zr-obp: Open bandit pipeline,'' 2021.
\newblock Accessed: 2024-07-12.

\bibitem{schlegel2022comparison}
K.~Schlegel, P.~Neubert, and P.~Protzel, ``A comparison of vector symbolic architectures,'' {\em Artificial Intelligence Review}, vol.~55, no.~6, pp.~4523--4555, 2022.

\bibitem{RW_review_4}
E.~Hassan, Y.~Halawani, B.~Mohammad, and H.~Saleh, ``Hyper-dimensional computing challenges and opportunities for ai applications,'' {\em IEEE Access}, vol.~10, pp.~97651--97664, 2021.

\bibitem{liu2018customized}
B.~Liu, T.~Yu, I.~Lane, and O.~Mengshoel, ``Customized nonlinear bandits for online response selection in neural conversation models,'' {\em Proceedings of the AAAI Conference on Artificial Intelligence}, vol.~32, Apr. 2018.

\bibitem{RW_robotic}
A.~Mitrokhin, P.~Sutor, C.~Ferm{\"u}ller, and Y.~Aloimonos, ``Learning sensorimotor control with neuromorphic sensors: Toward hyperdimensional active perception,'' {\em Science Robotics}, vol.~4, no.~30, p.~eaaw6736, 2019.

\bibitem{RW_Healthcare_1}
N.~Watkinson, D.~Devineni, V.~Joe, T.~Givargis, A.~Nicolau, and A.~Veidenbaum, ``Using hyperdimensional computing to extract features for the detection of type 2 diabetes,'' in {\em 2023 IEEE International Parallel and Distributed Processing Symposium Workshops (IPDPSW)}, pp.~149--156, IEEE, 2023.

\bibitem{RW_Textclassification_1}
F.~R. Najafabadi, A.~Rahimi, P.~Kanerva, and J.~M. Rabaey, ``Hyperdimensional computing for text classification,'' in {\em Design, automation test in Europe conference exhibition (DATE), University Booth}, pp.~1--1, 2016.

\bibitem{RW_Speech}
M.~Imani, D.~Kong, A.~Rahimi, and T.~Rosing, ``Voicehd: Hyperdimensional computing for efficient speech recognition,'' in {\em 2017 IEEE international conference on rebooting computing (ICRC)}, pp.~1--8, IEEE, 2017.

\bibitem{Rosato_RHDC}
A.~Rosato, M.~Panella, E.~Osipov, and D.~Kleyko, ``On effects of compression with hyperdimensional computing in distributed randomized neural networks,'' in {\em Advances in Computational Intelligence} (I.~Rojas, G.~Joya, and A.~Catal{\`a}, eds.), (Cham), pp.~155--167, Springer International Publishing, 2021.

\bibitem{RW_classification_review}
L.~Ge and K.~K. Parhi, ``Classification using hyperdimensional computing: A review,'' {\em IEEE Circuits and Systems Magazine}, vol.~20, no.~2, pp.~30--47, 2020.

\bibitem{HDCluster}
M.~Imani, Y.~Kim, T.~Worley, S.~Gupta, and T.~Rosing, ``Hdcluster: An accurate clustering using brain-inspired high-dimensional computing,'' in {\em 2019 Design, Automation and Test in Europe Conference and Exhibition (DATE)}, pp.~1591--1594, 2019.

\bibitem{RegHD}
A.~Hernandez-Cano, C.~Zhuo, X.~Yin, and M.~Imani, ``Reghd: Robust and efficient regression in hyper-dimensional learning system,'' in {\em 2021 58th ACM/IEEE Design Automation Conference (DAC)}, pp.~7--12, IEEE, 2021.

\bibitem{RW_DNA}
Y.~Kim, M.~Imani, N.~Moshiri, and T.~Rosing, ``Geniehd: Efficient dna pattern matching accelerator using hyperdimensional computing,'' in {\em 2020 Design, Automation \& Test in Europe Conference \& Exhibition (DATE)}, pp.~115--120, IEEE, 2020.

\bibitem{AeneasHDC_2024}
``Aeneashdc.'' \url{https://github.com/AeneasHDC/}, Jan. 2024.

\bibitem{HDCU}
F.~Author and S.~Author, ``Configurable hardware acceleration for hyperdimensional computing extension on risc-v,'' {\em TechRxiv}, 2024.

\bibitem{ni2022qhd}
Y.~Ni, D.~Abraham, M.~Issa, Y.~Kim, P.~Mercati, and M.~Imani, ``Qhd: A brain-inspired hyperdimensional reinforcement learning algorithm,'' {\em arXiv preprint}, 2022.

\bibitem{ni2022hdpg}
Y.~Ni, M.~Issa, D.~Abraham, M.~Imani, X.~Yin, and M.~Imani, ``Hdpg: Hyperdimensional policy-based reinforcement learning for continuous control,'' in {\em Proceedings of the 59th ACM/IEEE Design Automation Conference}, pp.~1141--1146, 2022.

\bibitem{Rachkovskiy_2005}
D.~A. Rachkovskiy, S.~V. Slipchenko, E.~M. Kussul, and T.~N. Baidyk, ``Sparse binary distributed encoding of scalars,'' {\em Journal of Automation and Information Sciences}, vol.~37, no.~6, pp.~12--23, 2005.

\bibitem{map}
R.~Gayler, ``Multiplicative binding, representation operators, and analogy [abstract of poster],'' {\em Advances in analogy research: Integration of theory and data from the cognitive, computational, and neural sciences, ed. K. Holyoak, D. Gentner \& B. Kokinov. New Bulgarian University. Available at: http://cogprints. org/502/.[RWG, FTS]}, 1998.

\bibitem{lattimore2020bandit}
T.~Lattimore and C.~Szepesv{\'a}ri, {\em Bandit algorithms}.
\newblock Cambridge University Press, 2020.

\bibitem{faster1}
C.-Y. Chang, Y.-C. Chuang, C.-T. Huang, and A.-Y. Wu, ``Recent progress and development of hyperdimensional computing (hdc) for edge intelligence,'' {\em IEEE Journal on Emerging and Selected Topics in Circuits and Systems}, vol.~13, no.~1, pp.~119--136, 2023.

\bibitem{faster2}
A.~Rahimi, S.~Benatti, P.~Kanerva, L.~Benini, and J.~M. Rabaey, ``Hyperdimensional biosignal processing: A case study for emg-based hand gesture recognition,'' in {\em 2016 IEEE International Conference on Rebooting Computing (ICRC)}, pp.~1--8, 2016.

\bibitem{LINUCBOPT}
M.~Angioli, M.~Barbirotta, A.~Cheikh, A.~Mastrandrea, F.~Menichelli, and M.~Olivieri, ``Efficient implementation of linearucb through algorithmic improvements and vector computing acceleration for embedded learning systems,'' 2025.

\bibitem{neubert2019introduction}
P.~Neubert, S.~Schubert, and P.~Protzel, ``An introduction to hyperdimensional computing for robotics,'' {\em KI-K{\"u}nstliche Intelligenz}, vol.~33, no.~4, pp.~319--330, 2019.

\end{thebibliography}
        
\end{document}